\documentclass[lettersize,journal]{IEEEtran}
\usepackage{amsmath,amsfonts}
\usepackage{algorithmic}
\usepackage{array}
\usepackage[caption=false,font=normalsize,labelfont=bf,textfont=rm]{subfig}
\usepackage{textcomp}
\usepackage{stfloats}
\usepackage{url}
\usepackage{verbatim}
\usepackage{graphicx}
\def\BibTeX{{\rm B\kern-.05em{\sc i\kern-.025em b}\kern-.08em
    T\kern-.1667em\lower.7ex\hbox{E}\kern-.125emX}}
\usepackage{balance}
\begin{document}
\title{Generative Principal Component Regression via Variational Inference}
\author{Austin Talbot, Corey J. Keller, David E. Carlson, Alex V. Kotlar%
\thanks{This work has been submitted to the IEEE for possible publication. Copyright may be transferred without notice, after which this version may no longer be accessible.}%
\thanks{This research was supported in part by the National Institute of Mental Health under award number R01MH126639 (AT, CJK), and a Burroughs Wellcome Fund Career Award for Medical Scientists (CJK). This work was also funded via a donation from Gates Ventures to the Goizueta ADRC at Emory University for the support of innovative work in the areas of brain imaging, genomics, and proteomics (AK). (Corresponding author: Austin Talbot).}%
\thanks{Austin Talbot is with Pillar Diagnostics Inc, Natick, MA 01760 USA (e-mail: talbota@pillarbiosci.com).}%
\thanks{Corey J. Keller is with 1) the Department of Psychiatry and Behavioral Sciences, Stanford University, Palo Alto, CA 94301 USA (e-mail: ckeller1@stanford.edu); 2) The Wu Tsai Neurosciences Institute, Stanford University, Stanford, CA, USA; 3) Veterans Affairs Palo Alto Healthcare System.}%
\thanks{David E. Carlson is with the Department of Civil and Environmental Engineering, Duke University, Durham, NC 27708 USA (e-mail: david.carlson@duke.edu).}%
\thanks{Alex V. Kotlar is with the Department of Biomedical Informatics, Emory University, Atlanta, GA 30322 USA (e-mail: alex.kotlar@emory.edu).}%
}

\markboth{Journal of \LaTeX\ Class Files,~Vol.~18, No.~9, September~2020}%
{How to Use the IEEEtran \LaTeX \ Templates}

\maketitle

\begin{abstract}

The ability to manipulate complex systems, such as the brain, to modify specific outcomes has far-reaching implications, particularly in the treatment of psychiatric disorders.  One approach to designing appropriate manipulations is to target key features of predictive models. While generative latent variable models, such as probabilistic principal component analysis (PPCA), is a powerful tool for identifying targets, they struggle incorporating information relevant to low-variance outcomes into the latent space. When stimulation targets are designed on the latent space in such a scenario, the intervention can be suboptimal with minimal efficacy. To address this problem, we develop a novel objective based on supervised variational autoencoders (SVAEs) that enforces such information is represented in the latent space. The novel objective can be used with linear models, such as PPCA, which we refer to as generative principal component regression (gPCR). We show in simulations that gPCR dramatically improves target selection in manipulation as compared to standard PCR and SVAEs. As part of these simulations, we develop a metric for detecting when relevant information is not properly incorporated into the loadings. We then show in two neural datasets related to stress and social behavior in which gPCR dramatically outperforms PCR in predictive performance and that SVAEs exhibit low incorporation of relevant information into the loadings. Overall, this work suggests that our method significantly improves target selection for manipulation using latent variable models over competitor inference schemes.

\end{abstract}


\section{Introduction}
\label{ssec:Introduction}
Latent variable models, particularly factor models, serve as a foundational tool across a broad spectrum of scientific disciplines. This ubiquity is unsurprising due to their ability to distill complex, high-dimensional data into a more manageable, low-dimensional form and the quick parameter convergence allowing the models to be inferred with relatively small sample sizes. They are used in astronomy to classify celestial bodies \cite{regier2015celeste} and in genomics to aid in the visualization and analysis of single-cell data \cite{lopez2018deep,wang2019data}. In the realm of social sciences, factor models uncover latent structures that can inform policy and planning decisions \cite{erosheva2007describing}. They are also heavily used in neuroscience \cite{cooper2019neuroimaging,porbadnigk2015extracting}, as their structure aligns with the idea of “networks” of relevant brain activity giving rise to the observed covariates \cite{bassett2017network}. In this field, the strong correlations between the covariates make sparsity in dimensionality a highly desirable model feature \cite{cunningham2014dimensionality}.

Beyond their use in exploratory data analysis, factor models are also used to develop hypotheses and targets for manipulations to modify an outcome or behavior associated with the data \cite{Hultman2018,fong2024principal,carlson2017dynamically}. Beyond the scientific goal of using manipulation to provide evidence of causality \cite{mague2022brain}, manipulations are critical in many clinical applications \cite{scangos2021closed}. Once the relationship between the factors and the outcome is known, targets can be chosen as influential covariates of the critical factors, as measured by the loadings. This approach has been used successfully to modify a diverse set of behaviors such as social activity \cite{mague2022brain}, aggression \cite{grossman2022brain}, and anxiety \cite{hughes2024widespread}. Unfortunately, while factor models excel in scientific interpretability, practical application of factor models in designing manipulations is quite difficult. The outcomes considered are commonly low variance signals and are easily overshadowed by more dominant high-variance components \cite{hahn2013partial}. Standard likelihood-based techniques may miss these subtle signals as they, by design, focus on explaining maximal variance. Because of this, fitting a predictive model subsequently to the generative model, such as in principal component regression (PCR) \cite{jolliffe1982note} and more broadly cutting the feedback \cite{mccandless2010cutting}, have performed poorly on prediction in comparison to solely predictive models

Addressing this problem often requires supervision, the incorporation of additional guiding signals—typically expressed as a loss function—that help steer the model towards learning representations that are specifically aligned with desired outcomes \cite{Mairal2009}. Supervised variational autoencoders (SVAEs) are a notable example of this approach \cite{kingma2014semi}. They employ an encoder-decoder structure to both compress the data into a latent space and reconstruct it, with the added supervision ensuring the encoded representations are pertinent to the outcome of interest. This approach ostensibly combines the best aspects of both generative and predictive models; the generative component adds to scientific interpretability and regularizes the supervision loss while the supervision ensures that the learned space is relevant to the outcome \cite{talbot2023estimating}.

However, recent work has shown that SVAEs possess a critical flaw when the loadings are used to design manipulations; the supervision loss “drags” the encoder away from the generative posterior, the distribution of the latent variables conditioned on the covariates, as defined by the loadings \cite{tu2022supervising}.  In other words, the latent variables implied solely by the generative model are different than the latent variables inferred by the full SVAE, and the generative latent variables can be dramatically worse for predicting the outcomes of interest. The discrepancy between the encoder and generative model is highly undesirable, as it means that manipulations based on the loadings may not modify the predictive space as desired or with dramatically reduced efficacy. This property had escaped detection as the use of the generative arm of the SVAE for target selection is a more recent application and less frequent.

In this paper, we develop a novel inference algorithm to address the issue of incorporating predictive information in generative models. This algorithm is straightforward to implement in linear models, which we term generative principal component regression (gPCR) that yields dramatically improved predictive performance from the latent variables implied by the generative model.  This is accomplished by using the SVAE objective but replacing the encoder with the generative posterior. This objective can be viewed as a solution to three separate problems: (1) inferring a linear predictive model with sparsity in dimensionality as opposed to covariates, (2) inferring a factor model relevant to an outcome of interest, and (3) eliminating the discrepancy between the encoder and decoder in SVAEs to improve experimental design, in this case for manipulation target selection. In addition, we also empirically demonstrate the problems caused by the encoder/decoder discrepancy in SVAEs, as we are able to directly compare the SVAE encoder with the generative posterior rather than relying on indirect evidence for the discrepancy. We evaluate our method on two neuroscience applications, one detecting the electrophysiology associated with stress and the other associated with social behavior and show that our method dramatically improves upon PCR and can match or exceed the performance of traditional predictive models. Finally, we show in synthetic data that our model provides superior identification of manipulation targets. Furthermore, we show that SVAEs exhibit similar behavior in the two neuroscience datasets, suggesting that this limitation is a real phenomenon rather than a theoretical concern and that our approach is a major advance in addressing this problem.

The contents of this paper are as follows: in Section \ref{ssec:RelatedWork} we summarize relevant work that either inspired our method or seeks to address this problem. In Section \ref{ssec:Methods} we derive our novel inference method and discuss its properties. In Section \ref{ssec:Synthetic}, we provide an illustrative example using synthetic data demonstrating how gPCR improves upon PCR for predictive ability and SVAEs for target selection. In Section \ref{ssec:Results} we demonstrate our inference algorithm's efficacy on multiple neuroscience datasets, along with illustrating the deficiencies of the commonly used SVAE. Finally, in Section \ref{ssec:Conclusion} we provide some brief remarks and potential future directions of this work. All models are implemented in the publicly available Bystro github repository \url{https://github.com/bystrogenomics/bystro} and all code required to reproduce the figures is located at \url{https://github.com/bystrogenomics/bystro-science}.

\section{Related Work}
\label{ssec:RelatedWork}
There are several areas of active research related to this work. First has been work on improving the predictive ability of latent variable models. One of the initial methods used thresholding to select the most predictive covariates \cite{Bair2006} and using these features for principal component regression. While effective, this has the undesirable impact of not including all covariates in the generative model, which is often undesirable scientifically. Other alternatives focus on making the generative model to reduce the impact of misspecification, introducing extra latent variables in partial least squares \cite{giessing2007fmri}, canonical correlation analysis \cite{zhuang2020technical}, or Bayesian nonparametric models \cite{bhattacharya2011sparse}. However, this additional flexibility often fails to improve predictive performance, which leads to methods for explicitly incorporating the auxiliary information \cite{mccandless2010cutting,Mairal2009,bernardo2003bayesian}, \cite{Mairal2009}. However, these methods also have struggled to properly incorporate information into the latent space \cite{talbot2023estimating}.

Another area of relevant research is to alter the use pseudolikelihoods used commonly in Markov random fields as opposed to traditional likelihoods \cite{besag1975statistical}. These methods replace the joint likelihood of the observed covariates with conditional likelihoods of each of the covariates conditioned on the remaining values, which avoids evaluating a computationally-intractable normalization constant. While superficially similar to gPCR, there are two critical differences. First, gPCR includes a joint likelihood of the remaining covariates making guarantees for likelihood-based inference still applicable. Second, and more importantly, gPCR upweights a specific conditional distribution of interest to improve predictive performance on the auxiliary variable.

Finally, recent developments in variational inference are also relevant to our work. Variational autoencoders allow for tractable inference on a wide variety of models \cite{kingma2013auto} by optimizing a lower bound on the likelihood \cite{blei2017variational}. This can be done by using a neural network “encoder” to approximate the generative posterior then using sampled values from the encoder to evaluate the generative model. This objective can be easily minimized using stochastic methods \cite{bottou2010large}, allowing for usage with large datasets using complex models. Furthermore, automatic differentiation in modern packages such as Pytorch allow for such models to be easily implemented.

\section{Deriving the Generative PCR Objective}
\label{ssec:Methods}
We start by defining notation. We are given demeaned samples $\{x_i\}_{i=1:N}\in\mathbb{R}^p$  and associated outcomes $\{y_i\}_{i=1:N}\in\mathcal{Y}$. Our objective is two-fold: we wish to develop a generative model with parameters $\theta$ to model $x$ and we would like this generative model to encode information about $y$. After specifying a prior $p_\theta(z)$  on the latent variables and the distribution of $x$ conditioned on $z$, $p_\theta(x|z)$, we obtain a model for $x$ as $p_\theta(x)=\in p_\theta(x|z)p_\theta(z)dz$. A natural and common way to model $y$ in this framework is to specify $p_\theta(y|z)$ and assume conditional independence between $x$ and $y$ \cite{yu2006supervised}. 

In this work, when developing practical inference methods, we will limit ourselves to linear models. That is, we assume that
\begin{align}
   p_\theta(z)&=N(0,I_L),\\
   p_\theta(x | z)&=N(Wz,\Lambda),
\end{align}
where $W\in\mathbb{R}^{p\times L}$ and $\Lambda$ is a diagonal matrix. This formulation corresponds to probabilistic PCA in the special case that $\Lambda=\sigma^2I$. However, we do not place any limitations on $p_\theta(y|z)$. Given the widespread use of linear models in a variety of scientific disciplines, this work yields a widely-applicable model \cite{greenacre2022principal,yeung2001principal,zhang2020fast}.

\subsection{Emphasizing the Desired Predictive Distribution}

Many of the difficulties in modern predictive tasks are due to the high dimensionality of $x$ resulting in difficult inference for $\theta$. One might assume that latent variable models would inherently possess superior performance and would be optimal under a perfectly specified model, as $p_\theta(y|z)$ is a low-dimensional distribution. However, when the numbers of samples dramatically exceed the numbers of parameters, solely predictive models tend to predict better in practice. The reason for the performance gap is simple: in the classical regime the parameters of a generative model that are effective at regularization are incredibly restrictive. This combined with the fact that the total variance in the high-dimensional $x$ is substantially larger than the variance in y this means that even minor misspecification in the generative model encourages the model to sacrifice $p_\theta(y|x)$  in favor of $p_\theta(x)$ under likelihood-based inference \cite{hahn2013partial}. To restate, even simple types of model misspecification, such as underestimating the true latent dimensionality, will dramatically degrade performance if y is correlated with the lower variance variables. 

A natural method to address this issue is to simply upweight the desired conditional distribution and maximize the modified objective. Such an objective (suppressing penalization terms or priors on $\theta$  for clarity) is 
\begin{equation}\label{eq:weightedconditional}
    \max_{\theta}\sum_{i=1}^N \log p_\theta(x_i) + \mu \log p_\theta(y_i|x_i),
\end{equation}
where $\mu$ is the tuning parameter controlling the emphasis on the predictive distribution of $y$. A value of $\mu=1$ corresponds to the standard maximum likelihood objective of the joint distribution, while larger values of $\mu$ correspond to an increasing emphasis on the specific conditional distribution. We can see from this that in most practical applications, $\mu$ will have to be very large, as $\log p_\theta(x)$ will be very large relative to $\log p_\theta(y|x)$. The objective above can be obtained rigorously as a Lagrangian relaxation \cite{lemarechal2001lagrangian} of maximizing the generative log likelihood with a constraint on the predictive distribution. Alternatively, this approach can be viewed as tempering the predictive distribution \cite{kapoor2022uncertainty} to increase its relative importance.

\subsection{Introducing a Targeted Variational Lower Bound}

Unfortunately, while the term $\log p_\theta(x)$  in (\ref{eq:weightedconditional}) has an analytic form in linear models, the term $\log p_\theta(y|x)=\int p_\theta(y|z)p_\theta(z|x)dz$ does not unless $y$ is also Gaussian. This is suboptimal for many classification applications, where logistic \cite{agresti2012categorical} or probit losses \cite{cutler2023quantitative} are desirable both theoretical and practical reasons. However, in this work we develop a second novel objective that eliminates this constraint using the same methods as variational autoencoders. This allows the use of any predictive loss or distribution to ensure a phenotypically relevant latent space.

To do this, we will make use of the following decomposition of the log likelihood, 
\begin{equation}\label{eq:kl_lower}
    \log p_\theta(x)=-D_{KL}(p_\theta(z|x)|p_\theta(z)) + E_{p_\theta(z|x)}[\log p_\theta(x|z)]
\end{equation}
When $p_\theta(z|x)$ is replaced by a density $q_\phi(z|x)$ with parameters $\phi$, Equation (\ref{eq:kl_lower}) becomes the classic evidence lower bound used in variational inference \cite{blei2017variational}. We can use this decomposition, combined with the conditional independence of $x$ and $y$ given $z$ to rewrite the maximum likelihood objective as
\begin{multline}\label{eq:lower_bound}
    \max_{\theta}\sum_{i=1}^N \log p_\theta(x_i,y_i) = \\
    \max_{\theta}\sum_{i=1}^N -D_{KL}(p_\theta(z|x_i,y_i)|p_\theta(z)) + \\E_{p_\theta(z|x_i,y_i)}[\log p_\theta(x_i|z) + \log p_\theta(y_i|z)]
\end{multline}
This looks similar to (\ref{eq:weightedconditional}), as we now have separated the joint distribution into a reconstruction term and a predictive term, with an additional term quantifying the divergence between the posterior on the latent variables and the prior. 

At this point we introduce a variational approximation and replace $p_\theta(z|x_i,y_i)$ with $p_\theta(z|x_i)$ and the resulting objective functions as a lower bound on the likelihood. While we will elaborate further below, the reason we use $p_\theta(z|x_i)$ rather than a more flexible $q_\phi(z|x_i)$ with new parameters $\phi$ is to ensure any predictive information relevant to $y$ in the latent space by necessity is contained in the loadings. This variational approximation is a lower bound on the likelihood, as we are omitting the information obtained from y on the latent space. With this substitution we can recombine the first two terms and weight the third term to obtain our robust variational objective
\begin{equation}\label{eq:gpca}
    \max_{\theta}\sum_{i=1}^N p_\theta(x_i) + \mu E_{p_\theta(z|x_i)}[ \log p_\theta(y_i|z)]
\end{equation}
This variational objective is almost identical to (\ref{eq:weightedconditional}), as it leaves the generative likelihood unaltered. However, it does not require integrating out z which makes it compatible with the reparameterization trick used in variational autoencoders. This form also provides an intuitive justification for the variational lower bound. If the supervision term were $E_{p_\theta(z|x_i,y_i)}[\log p_\theta(y_i|z)]$, the model would simply rely on $y_i$ to infer a relevant latent space rather than ensuring that the latent space is relevant even absent knowledge of the outcome, resulting in a large discrepancy between $p_\theta (z|x,y)$ and $p_\theta (z|x)$. 
This formulation ensures that $p_\theta (z|x)$, and by extension $p_\theta (y|x)=\int p_\theta (y|z) p_\theta (z|x)dz$, is prioritized.

\subsection{Inference Via Gradient Descent}

Nothing in the previous section requires that $p_\theta (x)$  be a linear model. However, we have limited our consideration to linear models as our practical inference scheme depends on both $p_\theta (z|x)$ and $-D_(KL ) (p_\theta (z|x,y)|p_\theta (z))$ to be analytic in (\ref{eq:gpca}).
If these quantities are available, inference can be performed using the reparameterization trick from standard VAEs \cite{kingma2013auto} however, with the parameters of the encoder defined by the generative model. Because of this, we restrict (5) to be a linear model, the gPCR objective.

This novel objective is straightforward to optimize using gradient descent-based methods using the same techniques used for variational autoencoders. However, unlike traditional variational autoencoders, we found that batch training yielded superior performance to stochastic methods. A potential explanation for this behavior is that the combination of a simple architecture, an analytic generative likelihood, and the lack of a separate encoder makes the objective substantially better behaved. Thus, rather than providing necessary regularization, stochastic methods instead result in a slower convergence rate to a good local optimum. We also found that gradient descent with momentum substantially outperformed more modern methods such as Adam \cite{kingma2014adam}. 

A final benefit of our formulation in linear models is that inference has the same computational complexity as standard linear regression. While matrix inversion (required for evaluating the Gaussian likelihood) generally scales as $\mathcal{O}(p^3 )$. By exploiting the Sherman-Woodbury matrix identity, we can reduce the computational cost to $\mathcal{O}(L^2 p)$ while maintaining the ability to propagate gradients. When $p$ is substantially larger than $L$, as is commonly the case in latent variable models, the $L^2$ term is nearly insignificant. Thus, from a computational point of view, the model described here has no drawbacks compared to any other version of regression lacking a closed-form solution (such as LASSO).

\section{Synthetic Results}
\label{ssec:Synthetic}
We now provide an in-silico demonstration how gPCR dramatically improves on PCR and potentially improves experimental manipulation efficacy by eliminating the encoder/decoder discrepancy present in SVAEs. Let the data generation mechanism be 
\begin{align}\label{eq:simulation}
p(z)&=N(0,\Lambda),\\
p(x|z)&=N(Wz,\sigma^2 I),\\
p(y^*|z)&=N(z_1,\tau),\\
y&=1_{y^*>0},
\end{align}
where $\Lambda$ is a diagonal matrix of ones except in the first entry which is substantially smaller than $1$ and $z_1$ denotes the first element in $z$. In other words, information about y is encoded in the lowest variance component. In this simulation, we set $p=440$, $L=10$, and $\sigma^2=1$ with a sample size of 2000. For ease of visualization, $W_1$ was $1$ for the first $40$ covariates and $0$ for the remainder. The remaining factors were generated as $W_{ij}\sim N(0,1)$, with no constraint on orthogonality. This lack of orthogonality was chosen as it allows for the encoder to perform “double-duty” due to the overlap between a high-variance component with no predictive ability and a low-variance highly predictive component. We then fit three models, PCR with logistic regression, an SVAE, and a model using our novel objective, all with 5 latent variables representing the common situation where the number of estimated components is fewer than the true number of components. 

In the SVAE, we used an affine encoder $q_\phi(z|x)=N(Ax,D)$, corresponding to the standard VAE setup of a separate parameterization for the mean and a diagonal covariance $D$. This simple encoder is not restrictive in this situation, as the true posterior $p_\theta(z|x)=N((\sigma^2I+W^TW)^{-1}W^T,I-(W^TW)/\sigma^2)$ is also Gaussian. Furthermore, we induce sparsity by supervising only the first latent variable, aligning with previous work \cite{Gallagher2017}. This choice of sparsity enhances interpretability as the loadings of the supervised factor are a scaled version of the predictive coefficients. This synthetic formulation has two enormously beneficial properties; (1) we can directly evaluate the impact of separating the encoder from the decoder without concerns about encoder capacity that would occur in deeper models and (2) we can directly evaluate the discrepancy between $q_\phi(z|x)$ and $p_\theta(z|x)$. We choose the correlation between the posterior mean and the encoder mean as our measure of similarity.

\begin{figure*}[!t]
\centering
\includegraphics[width=7.0in]{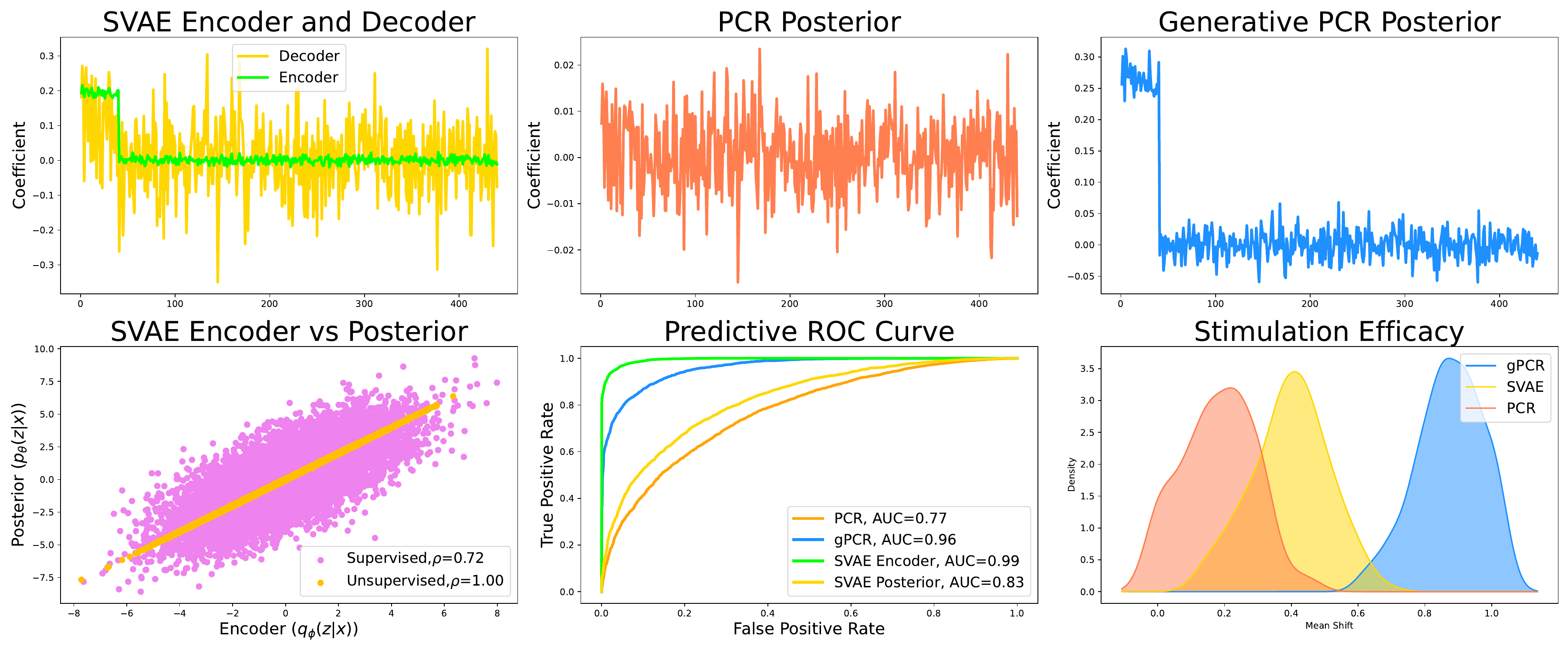}
\caption{The top left plot shows the encoder and decoder of the SVAE, the top middle shows the predictive coefficients as detected by PCR, and the top right shows the loadings of our robust model. In the bottom row we plot the encoder mean versus the posterior mean of an SVAE for the supervised and an unsupervised factor. In the middle we plot the predictive ability of the different models via an ROC curve. Finally, the bottom right shows the distribution of stimulation efficacies based on the different models.}
\label{fig1}
\end{figure*}

The different supervised components for all models are shown in the first row of figure \ref{fig1}. On the left we plot both the encoder and decoder in the SVAE, while the middle shows the coefficients of the learned linear model using PCR, which is a composition of the linear transformation and the subsequent regression coefficient, while on the right we show the loadings of gPCR, which is a scaled version of the predictive coefficients. We can see that the SVAE encoder and decoder differ dramatically. The encoder clearly detects that the first $40$ covariates are relevant to the outcome, the decoder is less clear. Certainly, the first $40$ covariates are highly influential but there are a substantial number of nonzero loadings among the remaining irrelevant coefficients. The decoder has created a superposition of several networks, while the encoder almost exclusively focuses on the predictive information the decoder becomes a superposition of networks; it explains the variance in both the supervised network and information contained in some of the remaining non-orthogonal networks. The PCR objective has captured minimal relevant information to the objective, which is unsurprising as the largest networks have minimal overlap with the supervised network. As a result, a large quantity of irrelevant high variance networks becomes incorporated in the resulting predictive model. Meanwhile, only gPCR is able to clearly separate the relevant coefficients from the irrelevant coefficients in the learned network.

We then show some of the signs that there is significant divergence between the encoder and decoder in an SVAE. First, as we visualize in the bottom left figure, the correlation between the posterior mean and encoder mean of the SVAE is dramatically reduced in the supervised factor as compared to an unsupervised factor. Given that supervision is isolated to a single factor, the encoder for the unsupervised networks is free to learn the optimal encoding for reconstruction loss. We can further detect this problem by a dramatic drop of predictive ability of the generative posterior as compared to the encoder as visualized in the middle plot. The encoder obtains almost perfect predictive ability with an AUC of 0.995. However, the predictions made by the latent variables inferred from the decoder (generative model) drop to $0.83$. While this is an improvement over standard PCR (AUC of $0.77$), it is dramatically degraded from the performance we would expect from the encoder. On the other hand, gPCR achieves an AUC of $0.96$, which is close to the predictive performance achievable by regression-based models.

Where gPCR truly shines is when the generative parameters are used to design stimulation procedures. We assume a causal relationship between x and y. We then create 100 distinct synthetic “stimulations” as shifting the mean of 10 randomly selected covariates by 1 from among the 50 largest covariates as measured by the generative parameters. We then examine the shift in $E[y^*]$ given each of the different stimulation techniques. This reflects the common biological situation where there are multiple candidates for stimulation given a network and the final protocol is chosen based on secondary criteria such as ease of access. The distribution of these stimulation procedures is shown in the bottom right. The protocols developed via PCR are minimally effective, which is unsurprising given that many of the influential covariates are independent under the true model. The SVAE is more effective, which we could see given that the supervision did alter the decoder to weight the initial $40$ covariates higher. However, target selection via gPCR is by far the most effective, with the average shift being 0.89, as opposed to $0.18$ for PCR and $0.41$ for the SVAE. As a result, we expect that stimulation targets in real datasets using gPCR should dramatically outperform SVAE and standard PCR.

\section{Imputation and Prediction in Neural Datasets}
\label{ssec:Results}

We demonstrate the advantages of our novel inference algorithm on two neuroscience datasets. The first dataset is publicly available \cite{carlson2023multi} and contains electrophysiological measurements of mice in a tail suspension experimental paradigm (TST). The objective of this experiment was to characterize electrophysiology in an animal model relevant to bipolar disorder. The recordings came from 26 mice, which were observed under various conditions—ranging from non-stressful (home cage) to highly stressful (tail suspension) —over a 20-minute period while continuously recording local voltages (LFPs) in 11 distinct brain regions. We segmented these recordings into 1-second intervals and estimated the spectral power in 1 Hz intervals from 1 to 56 Hz after performing preprocessing steps described in \cite{mague2022brain}, generating a total of 616 covariates. In this work we use the standardized log-transformed features, which is a common approach from signal processing \cite{clauset2009power,proakis2015digital}.

The second dataset (social) included electrophysiology from 28 mice recorded in 8 brain regions on multiple days. In each recording session, the mice were placed in a two-chambered social assay for 10 minutes. The mice were allowed to wander freely and in one chamber they were able to interact with another mouse (social interaction), while the other contained an inanimate object (non-social interaction). The location of the mouse was tracked during the entire recording and location was used as a proxy for social or non-social interaction. The initial objective of the experiment was to uncover the brain activity relevant to social interactions. The ultimate goal of developing stimulation targets to enhance social behavior, as currently medication struggles to treat social deficiencies in some disorders \cite{mccracken2021drug}. We used identical feature extraction steps used above to obtain 448 spectral power covariates.

\subsection{Regression: Imputing Unobserved Brain Activity}

The first application of our method is imputing dynamics of a missing brain region using the remaining regions in the TST dataset. This task is useful in its own right as missing data occurs for two common reasons. First, electrode failure is often observed, and while multiple electrodes are placed in each region, occasionally all electrodes fail or yield low-quality recordings resulting in no usable data from the specific region. Often, the data from these mice are not used, resulting in weeks of wasted effort. Second, data from multiple experiments are often used in a single study, for example, using mice from a different behavioral paradigm as a validation set for a specific hypothesis \cite{hughes2024widespread}. Depending on the priorities of the separate experiments, the recorded regions may not align, resulting in the need to infer the missing dynamics. For the purposes of this work, another advantage of a regression-based task is that it allows us to make direct comparisons with multiple alternative methods beyond principal component regression (PCR), namely partial least squares (PLS) and canonical correlation analysis (CCA). In this specific application, we are not limiting supervision to a single factor and instead use all latent factors for prediction. This reflects a difference in goal, rather than selection of stimulation targets we simply want to monitor activity in a potentially unmeasured region.

\begin{table}
\begin{center}
\caption{Region Imputation MSE}
\label{tab1}
\begin{tabular}{lcccc}
\hline
Method & Acumbens  & Thalamus & mSNC &  Hippocampus \\[5pt]
\hline
PCR & 0$\cdot$28$\pm$0$\cdot$01 & 0$\cdot$46$\pm$0$\cdot$01 & 0$\cdot$40$\pm$0$\cdot$01 & 0$\cdot$58$\pm$0$\cdot$01  \\
PLS & 0$\cdot$19$\pm$0$\cdot$01 & 0$\cdot$42$\pm$0$\cdot$01 & 0$\cdot$33$\pm$0$\cdot$01 & 0$\cdot$50$\pm$0$\cdot$02 \\
CCA & 0$\cdot$21$\pm$0$\cdot$01 & 0$\cdot$59$\pm$0$\cdot$01 & 0$\cdot$43$\pm$0$\cdot$01 & 0$\cdot$57$\pm$0$\cdot$02\\
\textbf{gPCR} & 0$\cdot$12$\pm$0$\cdot$01 & 0$\cdot$41$\pm$0$\cdot$02 & 0$\cdot$27$\pm$0$\cdot$01 & 0$\cdot$48$\pm$0$\cdot$03\\
ENet & 0$\cdot$07$\pm$0$\cdot$01 & 0$\cdot$39$\pm$0$\cdot$01 & 0$\cdot$19$\pm$0$\cdot$01 & 0$\cdot$50$\pm$0$\cdot$02 \\
\hline
\end{tabular}
\end{center}
\end{table}

In this experiment, we divided training and test sets by mouse to evaluate its performance in new animals \cite{walder2024electome} and repeated each experiment 50 time to obtain confidence intervals. In all dimension reduction models, 20 components were used. The results for several representative brain regions are shown in Table \ref{tab1}. We can see that Elastic Net outperforms traditional methods of PCA, PLS, and CCA universally. In some of these brain regions, such as acumbens or mSNC, the difference in performance is dramatic, with the MSE in acumbens being a third of the MSE in PCA and a quarter of the performance of CCA. Surprisingly, canonical correlation analysis, which is meant to address the issues outlined previously, underperforms the standard PCR in many regions. Our novel objective, in contrast, dramatically improves on the competitor methods, being close to linear regression in performance in all regions.

There are two important takeaways from these results. First, as previously mentioned, the capacity of generative models to make predictions is not the issue in their underperformance. Rather, it is a failure of likelihood-based inference methods to emphasize the desired characteristics of the model, namely good prediction. Second, the variational approximation does not impede predictive ability as we nearly match the performance of what is achievable with linear models. It is important to emphasize that unlike the predictive model, whose sole purpose is to impute unobserved dynamics, this is a full generative model that characterizes $p_\theta(x)$, and as such can be used for clustering \cite{yeung2001principal} and detect anomalies \cite{cai2021learned} and other tasks unable to be performed by the predictive model. Together, these results give us confidence that our inferred models in predictive tasks are achieving excellent performance where a consistent estimator is not available, such as the subsequent classification tasks. 

\subsection{Prediction: Stress Versus Nonstress Conditions}

We switch to classification tasks based on the original experimental justification. We no longer have an analytic form for $p_\theta(y|x)$, meaning that we cannot compare to PLS or CCA without changing from a logistic loss. However, PCR and logistic regression are still viable competitor methods. We start with the TST dataset and predict stress vs non-stress using the log spectral power features previously described. We compare our performance to PCR, $L_1$, $L_2$, and Elastic Net regression cross-validating over regularization strengths. We impose a sparseness penalty on the predictive coefficients of $p_\theta(y|z)$ to supervise only the first factor, similar to \cite{talbot2023estimating}.  In addition to improved biological interpretability (one network responsible for one behavior), it allows us to evaluate the effect that supervision has on the latent space.

We find that our supervised model almost matches the predictive performance of regression-based methods, with an AUC of $0.91\pm0.003$, as opposed to $0.94\pm0.003$ for $L_1$, $L_2$, and Elastic Net (EN) regression. However, this is a dramatic improvement over the performance of PCR, which has an AUC of 0.82±0.001. The predictive ability of this particular task is abnormally high, due to the dramatic differences between stressful and non-stressful conditions in mice. Because of this, even generative models are able to yield respectable predictive performance. However, even in these trivial tasks, predictive models yield superior performance.

\begin{figure}[!t]
\centering
\includegraphics[width=3.5in]{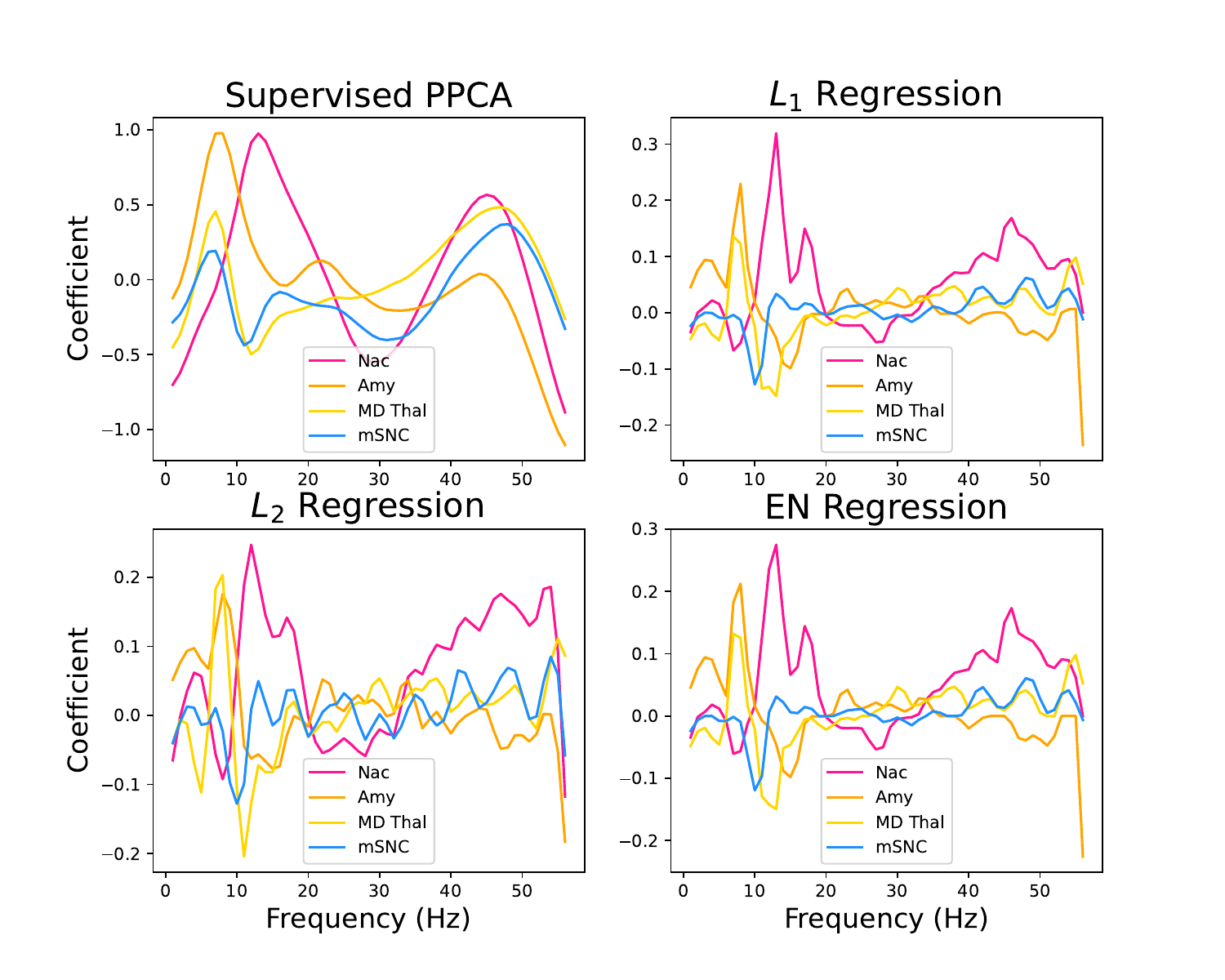}
\caption{\MakeLowercase{The predictive coefficients of stress versus nonstress conditions in four brain regions obtained via different regularization methods. Positive values indicate that spectral power is enhanced during stress while negative values indicate suppression.}}
\label{fig_coef_tst}
\end{figure}

While gPCR is unable to quite match the performance of regression-based models, it has dramatically more interpretable predictive coefficients, which are plotted in figure \ref{fig_coef_tst}. This plots the coefficients as a function of frequency in four representative brain regions. Positive coefficients indicate that spectral power is amplified in that band under stress, while negative coefficients indicate that power is suppressed. These features largely align between the different models, with an increase in power in NAC between 10 and 20 Hz and suppression of power in mSNC at 10 Hz. However, the gPCR model coefficients show dramatically smoother trajectories that we would expect based on the data. In any particular region, the effect that 10 Hz power should be largely similar to the effect of 11 Hz power. The jagged coefficients seen in the regression models are unrealistic and highlight the advantages of the latent variable viewpoint over a shrinkage viewpoint. 

\begin{table}
\begin{center}
\caption{Behavior Prediction AUCs}
\label{tab2}
\begin{tabular}{lcc}
\hline 
Method  & TST AUCs & Social AUCs \\
\hline 
PCR & 0$\cdot$819$\pm$0$\cdot$001 & 0$\cdot$51$\pm$0$\cdot$001 \\
$L_1$ Regression & 0$\cdot$937$\pm$0$\cdot$003 & 0$\cdot$55$\pm$0$\cdot$005 \\
$L_2$ Regression & 0$\cdot$937$\pm$0$\cdot$003 & 0$\cdot$57$\pm$0$\cdot$004  \\
Elastic Net Regression & 0$\cdot$937$\pm$0$\cdot$003 & 0$\cdot$57$\pm$0$\cdot$004 \\
\textbf{gPCR}  & 0$\cdot$913$\pm$0$\cdot$008 & 0$\cdot$57$\pm$0$\cdot$005 \\
\hline
\end{tabular}
\end{center}
\end{table}

\subsection{Prediction: Social Versus Nonsocial Interactions}

We now move on to an application that was the large motivation in developing these algorithms, distinguishing social from non-social interactions. Unsurprisingly, the differences between stress and nonstress conditions are substantially stronger than the differences in social vs nonsocial interaction, which is reflected in the weaker predictive performances observed in the latter experiment. This provides an ideal demonstration of the utility of gPCR, as now we are searching for relevant dynamics that are very weak. It is important to emphasize, however, that while the predictive relationships are certainly weaker, an AUC of $0.57$ was sufficient to design a stimulation protocol that successfully modified behavior \cite{mague2022brain}.

\begin{figure}[!t]
\centering
\includegraphics[width=3.5in]{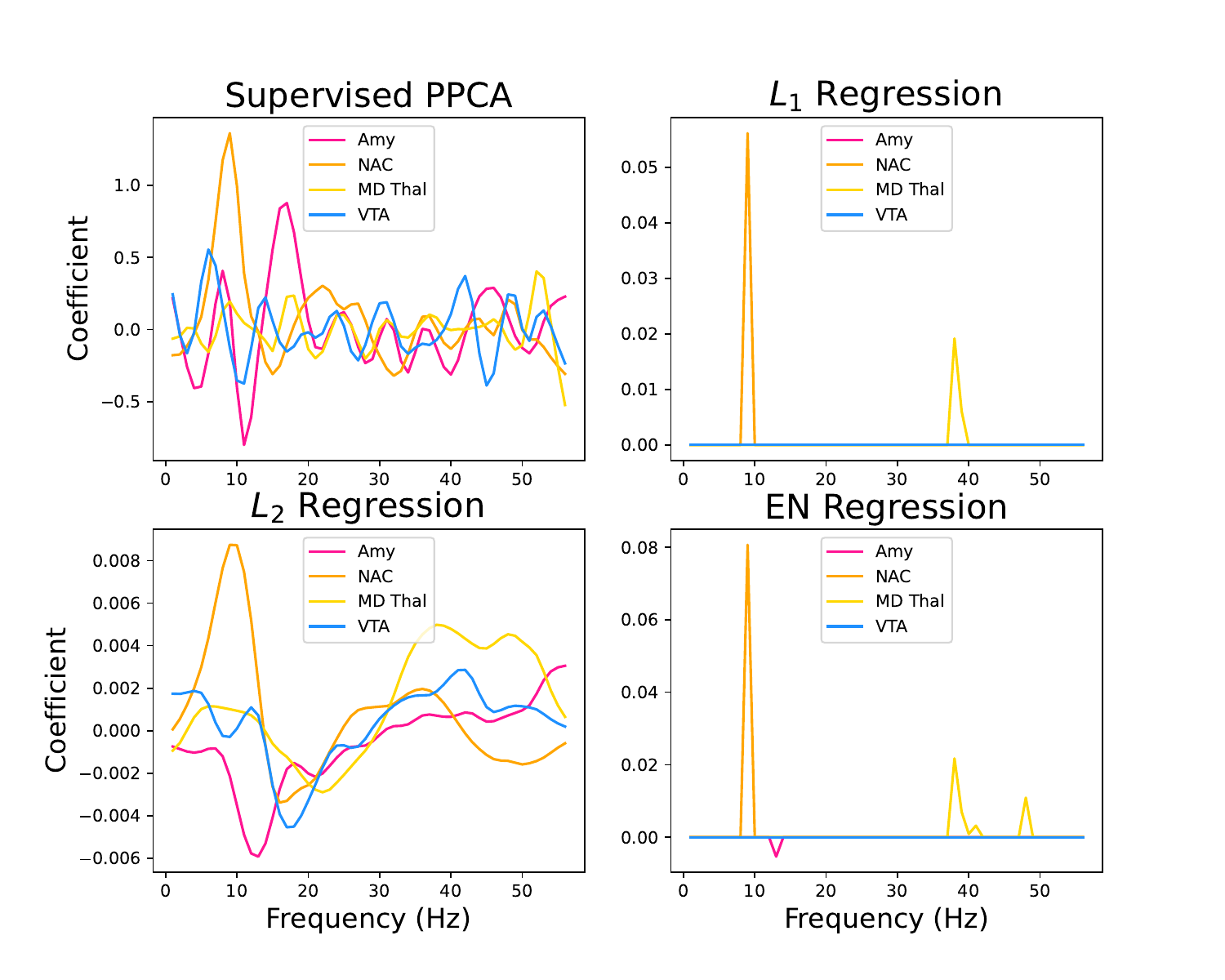}
\caption{The predictive coefficients of social versus nonsocial interactions. Positive values indicate power increases in social activity, negative values indicate suppression.}
\label{fig_coef_social}
\end{figure}

We found that $L_1$ regression yielded an AUC of $0.554\pm0.005$, EN regression had an AUC of $0.572\pm0.004$ and $L_2$ regression had an AUC of $0.575\pm0.004$. Incredibly, gPCR outperformed LASSO regression with an AUC of $0.57\pm0.005$ while matching the performance of $L_2$and EN regression. Given that gPCR must perform an additional task of reconstructing the data with a strong constraint on the predictive parameters, this was quite surprising. We found the origin of this discrepancy was overfitting on the part of the pure regression models. When the AUCs on the training set were examined, we found that $L_1$ regression outperformed gPCR (AUC of $0.63$ and $0.61$ respectively). Meanwhile, the PCR model had no predictive information with an AUC of $0.51\pm0.001$, even though the chosen dimensionality is large by the standards of neuroscience. Supervision in gPCR makes the difference between having no predictive ability and outperforming predictive models. This suggests several important conclusions. First, the posited latent network hypothesis is biologically realistic, as quantified in an objective comparison with predictive models that do not share this assumption. Second, it provides strong evidence in the efficacy of generative models to regularize predictive models. While all sparsity regularization was cross validated, for the latent variable models only a single set of parameters were used that had shown strong performance empirically, due purely to computational constraints.

This dataset also provides us an opportunity to evaluate the claim of improved parameter interpretability provided by a generative model as opposed to predictive models such as Lasso. While the previous task was sufficiently predictive that the penalization term was inconsequential, this task is sufficiently difficult that the penalization scheme makes a dramatic difference in the resulting coefficients, plotted in figure \ref{fig_coef_social}. LASSO and Elastic Net perform as they were designed, with the inherent sparsity assumption shrinking most of the coefficients to 0. Ridge regression did not shrink the coefficients to 0 and captured the expected smooth variation. However, the incredible amount of shrinkage required resulted in most of the coefficients being infinitesimal. The PCA model on the other hand had the large coefficients we would expect with relatively smooth variation we would expect. This is unsurprising, the requirement that the factor perform double duty of prediction and variance explanation in the electrophysiology requires that these coefficients be non-trivial and relatively smooth. This results in some dynamics missed in the other regression model to be captured in gPCR, which increases the variety of potential targets for stimulation. Given that some regions are more accessible than others, it is highly desirable that the model not run the risk of eliminating targets that are correlated but slightly less predictive in favor of a covariate that is difficult to modify.

\subsection{Exposing the Deficiencies of SVAE Loadings for Target Selection}

As our last contribution, we compare the results of fitting an SVAE as opposed to gPCR on the two neuroscience datasets. We use the same methodology in the second synthetic example to compare the posterior with the encoder and any discrepancies in the latent space. Unfortunately, due to the expense and time required to collect the data, performing a second stimulation protocol based on an SVAE that is hypothesized to perform worse is simply not viable. However, we can compare the other characteristics from the synthetic example that would suggest suboptimal loadings in an SVAE, namely lower correlations between the encoder and posterior means along with a drop in predictive accuracy when using the posterior mean for prediction.  

\begin{figure*}[!t]
\centering
\includegraphics[width=7.0in]{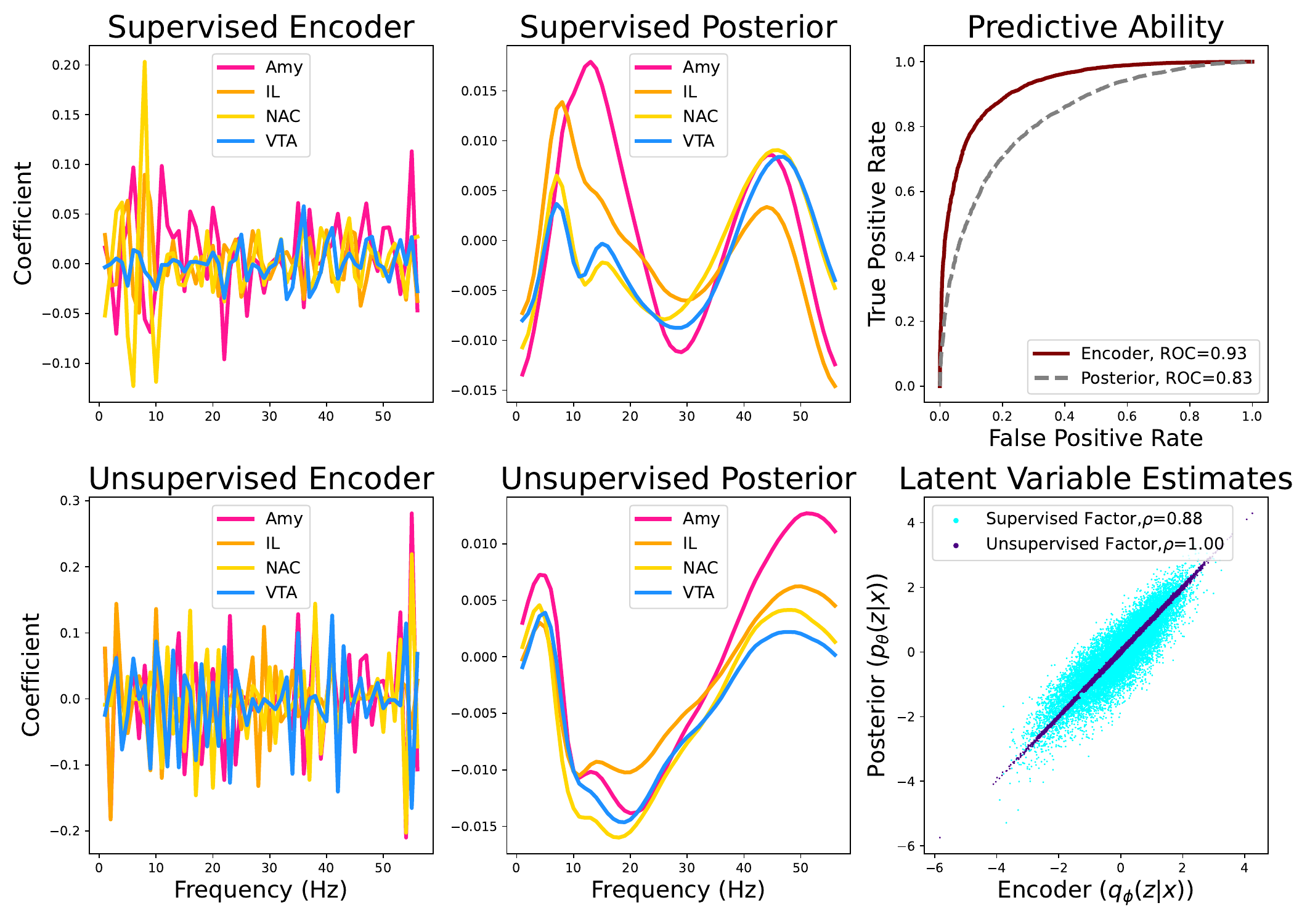}
\caption{This plot shows relevant quantities of the SVAE learned on the TST dataset. The top left plot shows the encoder and decoder of the SVAE, the top middle shows the predictive coefficients as detected by PCR, and the top right shows the loadings of our robust model. In the bottom row we plot the encoder mean versus the posterior mean of an SVAE for the supervised and an unsupervised factor. In the middle we plot the predictive ability of the different models via an ROC curve. Finally, the bottom right shows the distribution of stimulation efficacies based on the different models.}
\label{fig_tst_svae}
\end{figure*}

\begin{figure*}[!t]
\centering
\includegraphics[width=7.0in]{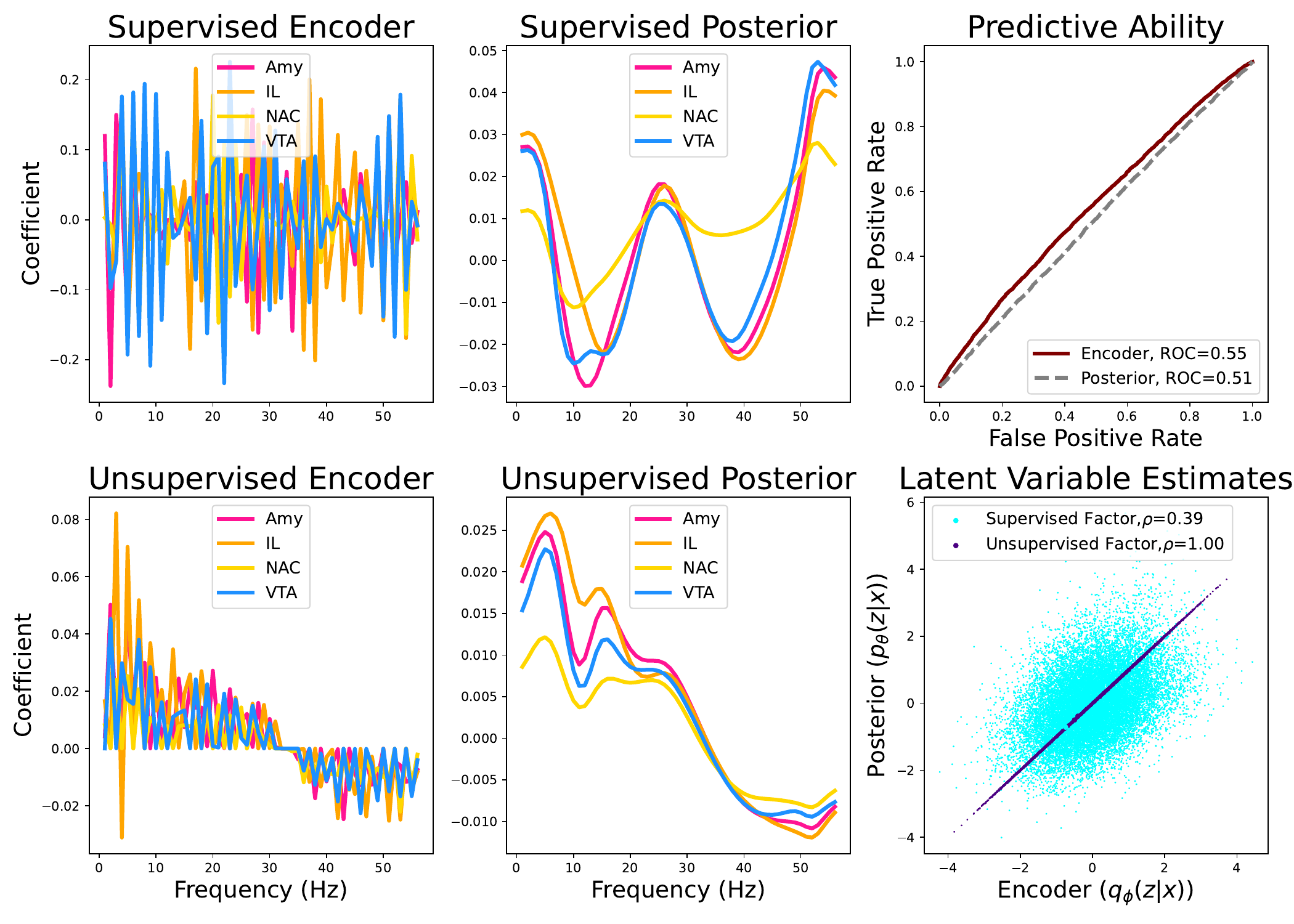}
\caption{This plot shows relevant quantities of the SVAE learned on the social preference dataset, with the interpretation matching figure \ref{fig_tst_svae}.}
\label{fig_sp_svae}
\end{figure*}

We show the relevant results from the model for the TST task in figure \ref{fig_tst_svae}. We can see dramatic differences between the learned encoder and the true posterior, as shown in the top left and top middle plots respectively. There is substantial jaggedness in the encoder that is not present in the decoder, which in part stems from additional regularization required to prevent overfitting on the predictive task. However, this is not the critical issue; instead, the critical flaw is that although the decoder shows power amplification in all regions at a wide range at $10$ and $50$ Hz, the encoder certainly does not support that conclusion. Furthermore, we can see a dramatic drop in predictive ability as shown by the ROC curves in the top right panel, with the encoder achieving an ROC of $0.93$ while the posterior has an AUC of $0.83$. We can see this discrepancy in the latent space as shown in the bottom right panel as the correlation between the two scores is only $\rho=0.88$. While there are some visual discrepancies between the encoder and decoder in the generative factors as shown by the bottom left and center panels, the latent states determined by the two methods correlate very strongly with $\rho=1.0$. In aggregate, these results are similar to those seen in the synthetic example. The discrepancies are even stronger in the social preference task as visualized in figure \ref{fig_sp_svae}. Here, the generative posterior has no predictive ability (AUC of $0.51$) and the estimates of the latent variables via the encoder have a substantially lower correlation from those provided via the generative model with $\rho=0.39$. Thus, while we are unable to perform the experiment in-vivo, these results strongly suggest that stimulation techniques based on gPCR would dramatically outperform those based on an SVAE, particularly in the social/non-social task.

\section{Conclusion}
\label{ssec:Conclusion}
Generative models, such as factor analysis, have many desirable properties, such as allowing for easy covariate imputation, a desirable scientific interpretation, and quick parameter convergence in terms of sample size. Unfortunately, they have been ignored in many predictive applications, as under mild model misspecification often results in poor predictive performance, unless the predictive task aligns with high-variance components. Here, we develop a novel inference objective that allows researchers to maintain all desirable properties of generative modeling, while ensuring that the latent variables are relevant to scientific questions. This is done by emphasizing a specific predictive distribution using a variational objective. This encourages that the model be predictive in terms of the generative parameters. We show that it is critical that this variational lower bound be obtained in terms of the generative posterior and that such an approach is competitive with traditional linear models in multiple applications. Furthermore, by avoiding the incorporation of a separate decoder, this approach forces the relevant information to be incorporated into the generative features, which is critical in many stimulation-based applications.

This work also leaves several promising avenues for extension. The most prominent is relaxing the requirement that $p_\theta(z|x)$ and $D_{KL}(p_\theta(z|x)|p_\theta(z))$ be analytic, allowing this technique to be used in a broader class of latent variable models. The second is further exploring why the SVAE approach struggles to incorporate the phenotypically relevant information into the generative parameters. Under the current model assumptions, the posterior mean is able to be properly represented by the linear encoder used in the variational lower bound, making such large impacts surprising. Finally, it would be helpful to demonstrate experimentally that stimulation based on gPCR outperforms competitor methods.

\balance

\section*{Acknowledgment}
\noindent ChatGPT-4o was used for editing and grammar enhancement throughtout the document.

\bibliographystyle{IEEEtran}
\bibliography{references}

\end{document}